\documentclass{bmvc2k}          
\title{DeepText: A Unified Framework for Text Proposal Generation and Text Detection in Natural Images}

\addauthor{Zhuoyao Zhong}{z.zhuoyao@mail.scut.sdu.cn}{1}
\addauthor{Lianwen Jin}{lianwen.jin@gmail.com}{1}
\addauthor{Shuye Zhang}{shuye.cheung@gmail.com}{1}
\addauthor{Ziyong Feng}{feng.ziyong@mail.scut.edu.cn}{1}

\addinstitution{
 School of Electronic and Information Engineering\\
 South China University of Technology\\
 Guangzhou, China
}

\runninghead{ Zhuoyao Zhong, Lianwen Jin, \emph{et al.}}{DeepText}

\usepackage{algorithm} 
\usepackage{algorithmic}

\begin{document}

\maketitle

\begin{abstract}
In this paper, we develop a novel unified framework called DeepText for text region proposal generation and text detection in natural images via a fully convolutional neural network (CNN). First, we propose the inception region proposal network (Inception-RPN) and design a set of text characteristic prior bounding boxes to achieve high word recall with only hundred level candidate proposals. Next, we present a powerful text detection network that embeds ambiguous text category (ATC) information and multi-level region-of-interest pooling (MLRP) for text and non-text classification and accurate localization. Finally, we apply an iterative bounding box voting scheme to pursue high recall in a complementary manner and introduce a filtering algorithm to retain the most suitable bounding box, while removing redundant inner and outer boxes for each text instance. Our approach achieves an F-measure of {\bf0.83} and {\bf0.85} on the ICDAR 2011 and 2013 robust text detection benchmarks, outperforming previous state-of-the-art results.
\end{abstract}

%------------------------------------------------------------------------- 
\section{Introduction}
\label{sec:introduction}
Text detection is a procedure that determines whether text is present in natural images and, if it is, where each text instance is located. Text in images provides rich and precise high-level semantic information, which is important for numerous potential applications such as scene understanding, image and video retrieval, and content-based recommendation systems. Consequently, text detection in natural scenes has attracted considerable attention in the computer vision and image understanding community \cite{Sliding-windows,DeepFeature,MSER1,MSER3,MSER4,CER,SWT,ICDAR2011, ICDAR2013,IJCV,TextFlow,CPN}. However, text detection in the wild is still a challenging and unsolved problem because of the following factors. First, a text image background is very complex and some region components such as signs, bricks, and grass are difficult to distinguish from text. Second, scene text can be diverse and usually exits in various colors, fonts, orientations, languages, and scales in natural images. Furthermore, there are highly confounding factors, such as non-uniform illumination, strong exposure, low contrast, blurring, low resolution, and occlusion, which pose hard challenges for the text detection task.

\begin{figure}
\centering
\includegraphics[width=100mm, height=48mm]{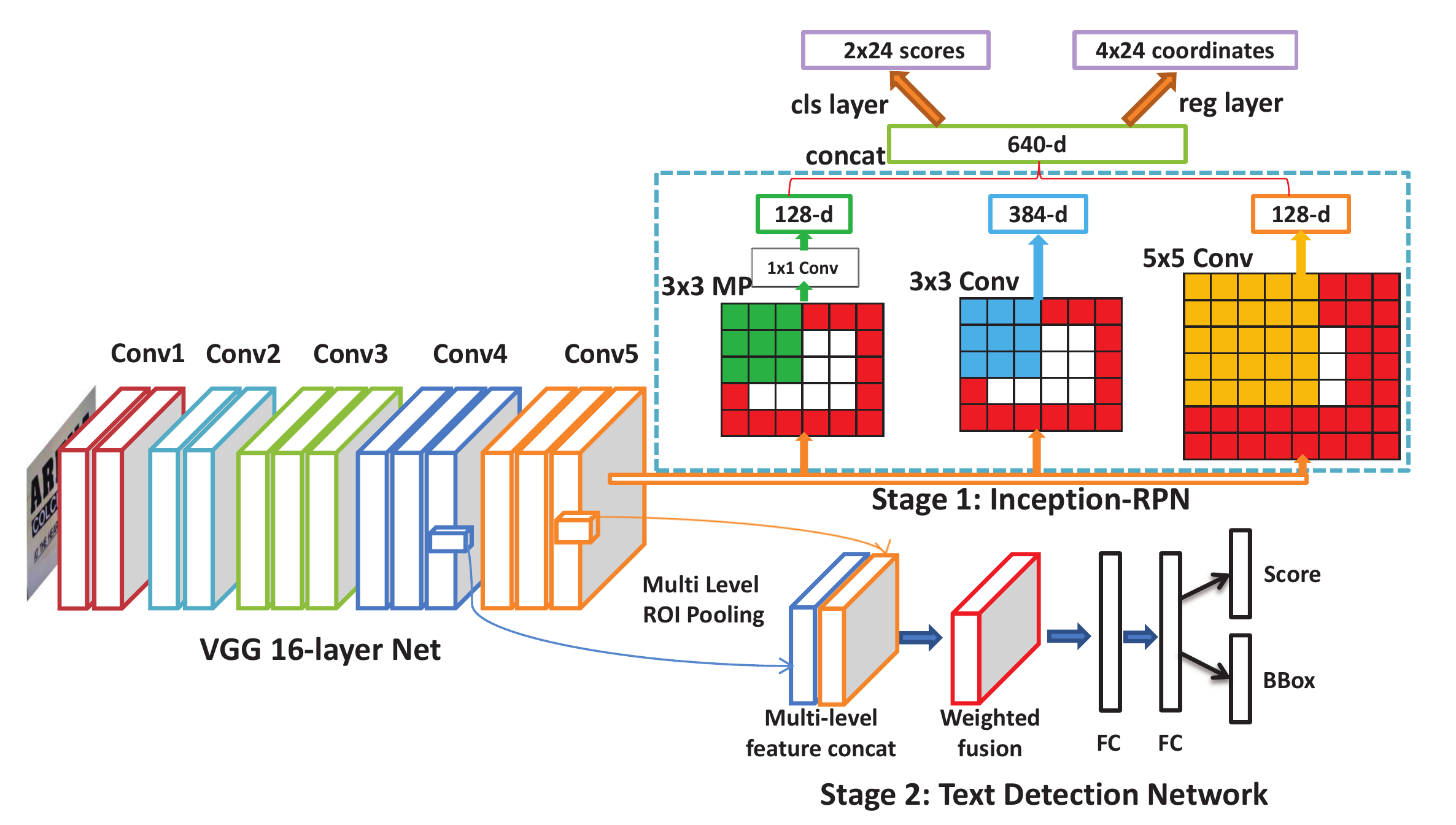}
\caption{
Pipeline architecture of DeepText. Our approach takes a natural image as input, generates hundreds of word region proposals via Inception-RPN (Stage 1), and then scores and refines each word proposal using the text detection network (Stage 2).
}
\label{fig:graph}
\end{figure}

In the last few decades, sliding window-based and connected component-based methods have become mainstream approaches to the text detection problem. Sliding window-based methods \cite{Sliding-windows,DeepFeature} use different ratios and scales of sliding windows to search for the presence of possible text positions in pyramid images, incurring a high computational cost. Connected component based methods, represented by maximally stable extremal regions (MSERs) \cite{MSER1,MSER3,MSER4,CER} and the stroke width transform (SWT) \cite{SWT}, extract character candidates and group them into word or text lines. In particular, previous approaches applying MSERs as the basic representation have achieved promising performance in the ICDAR 2011 and 2013 robust text detection competitions \cite{ICDAR2011,ICDAR2013}. However, MSERs focuses on low-level pixel operations and mainly accesses local character component information, which leads to poor performance in some challenging situations, such as multiple connected characters, segmented stroke characters, and non-uniform illumination, as mentioned in \cite{CPN}. Further, this bottom-up approach gives rise to sequential error accumulation in the total text detection pipeline, as stated in \cite{TextFlow}.

Rather than extract character candidates, Jaderberg \emph{et al.} \cite{IJCV} applied complementary region proposal methods called edge boxes (EB) \cite{EB} and aggregate channel feature (ACF) \cite{ACF} to perform word detection and acquired a high word recall with tens of thousands of word region proposals. They then employed HOG features and a random forest classifier to remove non-text region proposals and hence improve precision. Bounding box regression was also used for more accurate localization. Finally, using a large pre-trained convolutional neural network (CNN) to recognize the detected word-cropped images, they achieved superior text spotting and text-based image retrieval performance on several standard benchmarks.. 

Actually, the region proposal generation step in the generic object detection pipeline has attracted much interest. In recent studies, object detection models based on region proposal algorithms to hypothesize class-specific or class-agnostic object locations have achieved state-of-the-art detection performance \cite{RCNN,Fast-RCNN,SPPNet,m-RCNN}. However, standard region proposal algorithms such as selective search (SS) \cite{SS}, MCG \cite{MCG}, EB \cite{EB}, generate an extremely large number of region proposals. This leads to high recall, but burdens the follow-up classification and regression models and is also relatively time-consuming. In order to address these issues, Ren \emph{et al.} \cite{Faster-RCNN} proposed region proposal networks (RPNs), which computed region proposals with a deep fully CNN. They generated fewer region proposals, but achieved a promising recall rate under different overlap thresholds. Moreover, RPN and Fast R-CNN can be combined into a joint network and trained to share convolutional features. Owing to the above innovation, this approach achieved better object detection accuracy in less time than Fast R-CNN with SS \cite{Fast-RCNN} on PASCAL VOC 2007 and 2012.

In this paper, inspired by \cite{Faster-RCNN}, our motivation is to design a unified framework for text characteristic region proposal generation and text detection in natural images. In order to avoid the sequential error accumulation of bottom-up character candidate extraction strategies, we focus on word proposal generation. In contrast to previous region proposal methods that generate thousands of word region proposals, we are motivated to reduce this number to hundreds while maintaining a high word recall. To accomplish this, we propose the novel inception RPN (Inception-RPN) and design a set of text characteristic prior bounding boxes to hunt high-quality word region proposals. Subsequently, we present a powerful text detection network by incorporating extra ambiguous text category (ATC) information and multi-level region of interest (ROI) pooling into the optimization process. Finally, by means of some heuristic processing, including an iterative bounding box voting scheme and filtering algorithm to remove redundant boxes for each text instance, we achieve our high-performance text detection system, called DeepText. An overview of DeepText is shown in Fig. 1. Our contributions can be summarized by the following points.

(1) We propose inception-RPN, which applies multi-scale sliding windows over the top of convolutional feature maps and associates a set of text characteristic prior bounding boxes with each sliding position to generate word region proposals. The multi-scale sliding-window feature can retain local information as well as contextual information at the corresponding position, which helps to filter out non-text prior bounding boxes. Our Inception-RPN enables achieving a high recall with only hundreds of word region proposals.

(2) We introduce the additional ATC information and multi-level ROI pooling (MLRP) into the text detection network, which helps it to learn more discriminative information for distinguishing text from complex backgrounds.

(3) In order to make better use of intermediate models in the overall training process, we develop an iterative bounding box voting scheme, which obtains high word recall in a complementary manner. Besides, based on empirical observation, multiple inner boxes or outer boxes may simultaneously exist for one text instance. To tackle this problem, we use a filtering algorithm to keep the most suitable bounding box and remove the remainders.

(4) Our approach achieves an F-measure of 0.83 and 0.85 on the ICDAR 2011 and 2013 robust text detection benchmarks, respectively, outperforming the previous state-of-the-art results.

The remainder of this paper is set out as follows. The proposed methodology is described in detail in Section 2. Section 3 presents our experimental results and analysis. Finally, the conclusion is given in Section 4.

\section{Methodology}
\label{sec:methodology}

\subsection{Text region proposal generation}
Our inception-RPN method resembles the notion of RPN proposed in \cite{Faster-RCNN}, which takes a natural scene image and set of ground-truth bounding boxes that mark text regions as input and generates a manageable number of candidate word region proposals. To search for word region proposals, we slide an inception network over the top of convolutional feature maps (Conv5\_3) in the VGG16 model \cite{VGG16} and associate a set of text characteristic prior bounding boxes with each sliding position. The details are as follows.

\noindent{{\bf Text characteristic prior bounding box design.}} Our prior bounding boxes are similar to the anchor boxes defined in RPN. Taking text characteristics into consideration, for most word or text line instances, width is usually greater than height; in other words, their aspect ratios are usually less than one. Furthermore, most text regions are small in natural images.  Therefore, we empirically design four scales (32, 48, 64, and 80) and six aspect ratios (0.2, 0.5, 0.8, 1.0, 1.2, and 1.5), for a total of ${{k}=24}$ prior bounding boxes at each sliding position, which is suitable for text properties as well as incident situations. In the learning stage, we assign a positive label to a prior box that has an intersection over union (IoU) overlap greater than 0.5 with a ground-truth bounding box, while assigning a background label to a prior box with an IoU overlap less than 0.3 with any ground-truths.

\noindent{{\bf Inception-RPN.}} We design Inception-RPN, inspired by the idea of the inception module in GoogLeNet \cite{GoogLeNet}, which used flexible convolutional or pooling kernel filter sizes with a layer-by-layer structure to achieve local feature extraction. This method has proved to be robust for large-scale image classification. Our designed inception network consists of a ${3\times3}$ convolution, ${5\times5}$ convolution and ${3\times3}$ max pooling layers, which is fully connected to the corresponding spatial receptive fields of the input Conv5\_3 feature maps. That is, we apply ${3\times3}$ convolution,  ${5\times5}$ convolution and ${3\times3}$ max pooling to extract local featire representation over Conv5\_3 feature maps at each sliding position simultaneously. In addition, ${1\times1}$ convolution is employed on the top of ${3\times3}$ max pooling layer for dimension reduction. We then concatenate each part feature along the channel axis and a 640-d concatenated feature vector is fed into two sibling output layers: a classification layer that predicts textness score of the region and a regression layer that refines the text region location for each kind of prior bounding box at this sliding position. An illustration of Inception-RPN is shown in the top part of Fig. 1. Inception-RPN has the following advantages: (1) the multi-scale sliding-window feature can retain local information as well as contextual information thanks to its center restricted alignment at each sliding position, which helps to classify text and non-text prior bounding boxes, (2) the coexistence of convolution and pooling is effective for more abstract representative feature extraction, as addressed in \cite{GoogLeNet}, and (3) experiments shows that Inception-RPN substantially improves word recall at different IoU thresholds with the same number of word region proposals. 

Note that for a Conv5\_3 feature map of size ${{m}\times{n}}$, Inception-RPN generates ${{m}\times{n}\times24}$ prior bounding boxes as candidate word region proposals, some of which are redundant and highly overlap with others. Therefore, after each prior bounding box is scored and refined, we apply non-maximum suppression (NMS) \cite{NMS} with an IoU overlap threshold of 0.7 to retain the highest textness score bounding box and rapidly suppress the lower scoring boxes in the neighborhood. We next select the top-2000 candidate word region proposals for the text detection network.

\subsection{Text Detection}
\noindent{{\bf ATC incorporation}.} As in many previous works (e.g., \cite{Faster-RCNN}), a positive label is assigned to a proposal that has an IoU overlap greater than 0.5 with a ground truth bounding box, while a background label is assigned to a proposal that has an IoU overlap in the range ${\left[0.1,0.5\right)}$ with any ground-truths in the detection network. However, this method of proposal partitioning is unreasonable for text because a proposal with an IoU overlap in the interval ${\left[0.2,0.5\right)}$ may probably contain partial or extensive text information, as shown in Fig. 2. We note that promiscuous label information may confuse the learning of the text and non-text classification network. To tackle this issue, we refine this proposal label partition strategy to make it suitable for text classification. Hence, we assign a positive text label to a proposal that has an IoU overlap greater than 0.5 with a ground truth, while assigning an additional "ambiguous text" label to a proposal that has an IoU overlap with a ground truth bounding box in the range ${\left[0.2,0.5\right)}$. In addition, a background label is assigned to any proposal that has an IoU overlap of less than 0.2 with any ground-truths. We assume that more reasonable supervised information incorporation helps the classifier to learn more discriminative feature to distinguish text from complex and diverse backgrounds and filter out non-text region proposals.

\begin{figure}
\centering
\includegraphics[width=90mm,height=25mm]{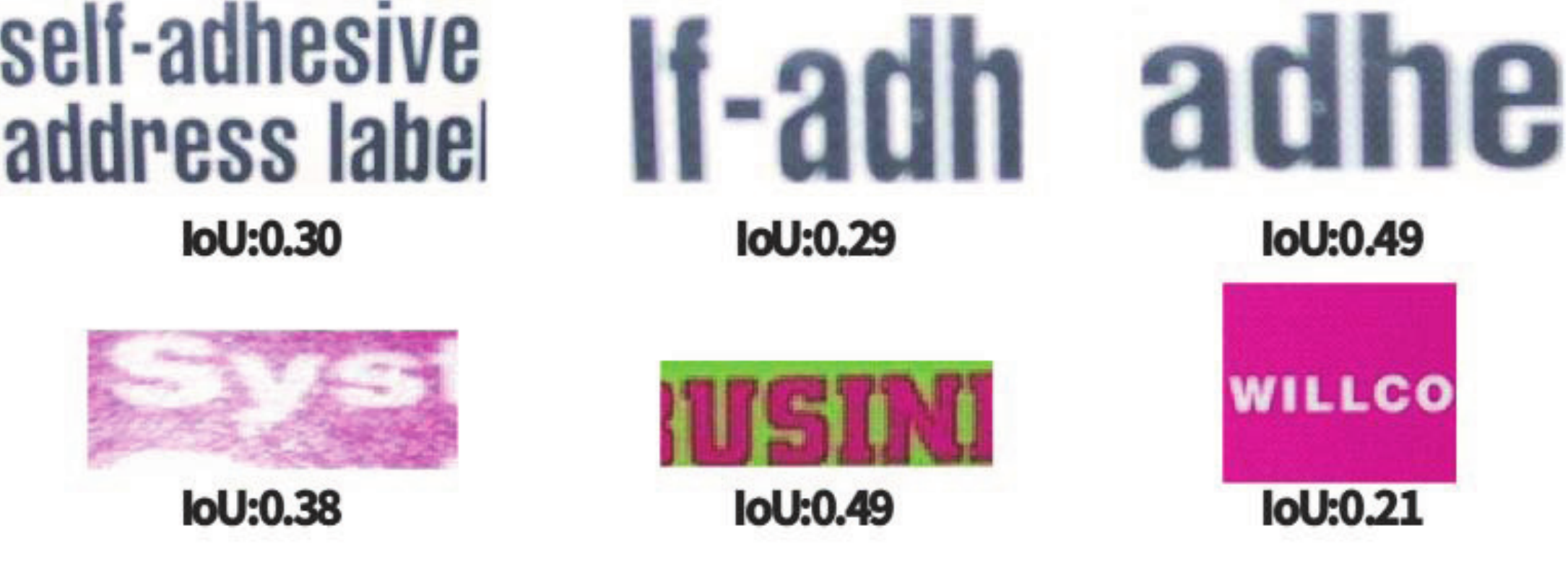}
\caption{
Example word region proposals with an IoU overlap within the interval ${\left[0.2,0.5\right)}$.
}
\label{fig:graph}
\end{figure}

\noindent{{\bf{MLRP.}}} The ROI pooling procedure performs adaptive max pooling and outputs a max-pooled feature with the original ${C}$ channels and spatial extents ${{H}\times{W}}$ for each bounding box.Previous state-of-the-art object detection models such as SPP-Net \cite{SPPNet}, fast-RCNN \cite{Fast-RCNN}, faster-RCNN \cite{Faster-RCNN}, all simply apply ROI pooling over the last convolutional layer (Conv5\_3) in the VGG16 model. However, to better utilize the multi-level convolutional features and enrich the discriminant information of each bounding box, we perform MLRP over the Conv4\_3 as well as Conv5\_3 convolutional feature maps of the VGG16 network and obtain two ${512\times{H}\times{W}}$ pooled feature (both ${{H}}$ and ${{W}}$ are set to 7 in practice). We apply channel concatenation on each pooled feature and encode concatenated feature with ${512\times1\times1}$ convolutional layer. The ${1\times1}$ convolutional layer is: (1) combines the multi-level pooled features and learns the fusion weights in the training process and (2) reduces the dimensions to match VGG16's first fully-connected layer. The multi-level weighted fusion feature is then accessed to the follow-up bounding box classification and regression model. An illustration of MLRP is depicted in the bottom half of Fig. 1.

\subsection{End-to-end learning optimization}
Both Inception-RPN and the text detection network have two sibling output layers: a classification layer and a regression layer. The difference between them is as follows: (1) For Inception-RPN, each kind of prior bounding box should be parameterized independently, so we need to predict all of the ${{k=24}}$ prior bounding boxes simultaneously. The classification layer outputs ${2k}$ scores textness scores that evaluate the probability of text or non-text for each proposal, while the regression layer outputs ${4k}$ values that encode the offsets of the refined bounding box. (2) For the text detection network, there are three output scores corresponding to the background, ambiguous text, and positive text categories and four bounding box regression offsets for each positive text proposal (only positive text region proposals access the bounding regression model). We minimize a multi-task loss function, as in \cite{RCNN}:
\begin{equation}
L(p,p^{*},t,t^{*})=L_{cls}(p,p^{*})+{\lambda}L_{reg}(t,t^{*}),
\end{equation}
where classification loss ${L_{cls}}$ is a softmax loss and ${p}$ and ${p^{*}}$ are given as the predicted and true labels, respectively. Regression loss ${L_{reg}}$  applies smooth-${\textup{L}_{1}}$ loss defined in \cite{Fast-RCNN}. Besides, ${t=\{t_{x},t_{y},t_{w},t_{h}\}}$ and ${t^{*}=\{t^{*}_{x},t^{*}_{y},t^{*}_{w},t^{*}_{h}\}}$ stand for predicted and ground-truth bounding box regression offset vector respectively, where ${t^{*}}$ is encoded as follows:
\begin{equation}
t^{*}_{x}=\frac{(G_{x}-P_{x})}{P_{w}},\quad t^{*}_{y}=\frac{(G_{y}-P_{y})}{P_{h}}, \quad
t^{*}_{w}=log(\frac{G_{w}}{P_{w}}), \quad
t^{*}_{h}=log(\frac{G_{h}}{P_{h}}). \quad
\end{equation}
Here, ${{P}=\{P_{x},P_{y},P_{w},P_{h}\}}$ and ${{G}=\{G_{x},G_{y},G_{w},G_{h}\}}$ denote the center coordinates (x-axis and y-axis), width, and height of proposal ${{P}}$ and ground-truth box ${{G}}$, respectively. Furthermore, ${\lambda}$ is a loss-balancing parameter, and we set ${\lambda=3}$ for Inception-RPN to bias it towards better box locations and ${\lambda=1}$ for text detection network.

In contrast to the proposed four-step training strategy to combine RPN and Fast-RCNN in \cite{Faster-RCNN}, we train our inception-RPN and text detection network in an end-to-end manner via back-propagation and stochastic gradient descent (SGD), as given in Algorithm 1. The shared convolutional layers are initialized by a pre-trained VGG16 model for imageNet classification \cite{VGG16}. All the weights of the new layers are initialized with a zero mean and a standard deviation of 0.01 Gaussian distribution. The base learning rate is 0.001 and is divided by 10 for each 40K mini-batch until convergence. We use a momentum of 0.9 and weight decay of 0.0005. All experiments were conducted in Caffe \cite{Caffe}.

\subsection{Heuristic processing}
\noindent{{\bf Iterative bounding box voting.}} In order to make better use of the intermediate models in the total training process, we propose an iterative bounding box voting scheme, which can be considered as a simplified version of the method mentioned in \cite{m-RCNN}. We use ${{D^{t}_{c}=\{B^{t}_{i,c},S^{t}_{i,c}\}}^{N_{c,t}}_{i=1}}$ to denote the set of $N_{c,t}$ detection candidates generated for specific positive text class ${c}$ in image ${I}$ on iteration ${t}$, where ${B^{t}_{i,c}}$ the \emph{i}-th bounding box and ${S^{t}_{i,c}}$ is the corresponding textness score. For ${t=1,...{T}}$, we merge each iteration detection candidate set together and generate ${D_{c}=\bigcup^{T}_{t=1}{D^{t}_{c}}}$ . We then adopt NMS \cite{NMS} on ${D_{c}}$ with an IoU overlap threshold of 0.3 to suppress low-scoring window boxes. In this way, we can obtain a high recall of text instances in a complementary manner and improve the performance of the text detection system.

\noindent{{\bf Filtering.}} Based on empirical observation, we note that even after NMS \cite{NMS} processing, multiple inner boxes or outer boxes may still exist for one text instance in the detection candidate set, which may severely harm the precision of the text detection system. To address this problem, we present a filtering algorithm that finds the inner and outer bounding boxes of each text instance in terms of coordinate position, preserves the bounding box with the highest textness score, and removes the others. Thus, we can remove redundant detection boxes and substantially improve precision.

\section{Experiments and Analysis}
\label{sec:experiments}

\begin{algorithm}[htb]
\caption{End-to-end optimization method for the DeepText training process.} 
\label{alg:Framwork}

\begin{algorithmic}[1] %1 controls mark number
\REQUIRE ~~\\
Set of training images with ground-truths: ${\{(I_{i},\{G_{i}\})\},...,(I_{N},\{G_{N}\}))}$; learning rate ${\eta(t)}$ ; samples number ${N_{*}=\{N_{b},N_{p},N_{a},N_{n}\}}$; iteration number ${t=0}$.
\ENSURE ~~\\   
Separate network parameters ${{\bf{W^{c}}},{\bf{W^{p}}},{\bf{W^{d}}}}$ for the shared convolutional layer, inception-RPN and text detection network. 
\STATE Randomly select one sample $(I_{i},\{G_{i}\})$ and produce prior bounding boxes classification labels and bounding box regression targets according to ${\{G_{i}\}}$;
\STATE Randomly sample ${N_{b}}$ positive and ${N_{b}}$ negative prior bounding box from ${\{G_{i}\}}$ to compute the loss function in equations (1);
\STATE Run backward propagation to obtain the gradient with respect to network parameters ${\nabla{\bf{W^{c}_{p}}},\nabla{\bf{W^{p}}}}$  and obtain the word proposal set ${\{P_{i}\}}$;
\STATE Adopt NMS with the setting IoU threshold on ${\{P_{i}\}}$ and select the top-\emph{k} ranked proposals to construct ${\{D_{i}\}}$ for Step 5;
\STATE Randomly sample ${N_{p}}$ positive text, ${N_{a}}$ ambiguous text and ${N_{n}}$ background word region proposals from ${\{D_{i}\}}$ to compute the loss function in equations (1);
\STATE Run backward propagation to obtain the gradient with respect to network parameters: ${\nabla{\bf{W^{c}_{d}}},\nabla{\bf{W^{d}}}}$;
\STATE update network parameters: ${{\bf{W^{c}}}={\bf{W^{c}}}-{\eta(t)\cdot{(\nabla{\bf{W^{c}_{p}}}+\nabla{\bf{W^{c}_{d}}})}}}$, ${{\bf{W^{p}}}={\bf{W^{p}}}-{\eta(t)\cdot{\nabla{\bf{W^{p}}}}}}$, ${{\bf{W^{d}}}={\bf{W^{d}}}-{\eta(t)\cdot{\nabla{\bf{W^{d}}}}}}$;
\STATE ${t=t+1}$, if the network has converged,output network parameters ${{\bf{W^{c}}},{\bf{W^{p}}},{\bf{W^{d}}}}$  and end the procedure; otherwise, return the Step 1.
\end{algorithmic}
\end{algorithm}

\subsection{Experiments Data}
The ICDAR 2011 dataset includes 229 and 255 images for training and testing, respectively, and there are 229 training and 233 testing images in the ICDAR 2013 dataset. Obviously, the number of training image is constrained to train a reasonable network. In order to increase the diversity and number of training samples, we collect an indoor database that consisted of 1,715 natural images for text detection and recognition from the Flickr website, which is publicly available online\footnote{\url{https://www.dropbox.com/s/06wfn5ugt5v3djs/SCUT\_FORU\_DB\_Release.rar?dl=0}} and free for research usage. In addition, we manually selected 2,028 images from the COCO-Text benchmark \cite{COCO}. Ultimately, we collected 4,072 training images in total.
\subsection{Evaluation of Inception-RPN}
In this section, we compare Inception-RPN with the text characteristic prior bounding boxes (Inception-RPN-TCPB) to state-of-the-art region proposal algorithms, such as SS \cite{SS}, EB \cite{EB} and standard RPN \cite{Faster-RCNN}. We compute the word recall rate of word region proposals at different IoU overlap thresholds with  ground-truth bounding boxes on the ICDAR 2013 testing set, which includes 1095 word-level annotated text regions. In Fig. 3, we show the results of using \emph{{N}}= 100, 300, 500 word region proposals, where the \emph{{N}} proposals are the top-\emph{{N}} scoring word region proposals ranked in term of these methods. The plots demonstrate that our Inception-RPN-TCPB considerably outperforms standard RPN by 8\%-10\% and is superior to SS and EB with a notable improvement when the number of word region proposals drops from 500 to 100. Therefore, our proposed Inception-RPN-TCPB is capable of achieving a high recall of nearly 90\% with only hundreds of word region proposals. Moreover, the recall rate of using 300 word region proposals approximates that of using 500 word region proposals, so we simply use the top-300 word region proposals for the text detection network at test time.

\subsection{Analysis of text detection network}
In this section, we investigate the effect of ATC incorporation and MLRP on the text detection network. First, we use our proposed Inception-RPN-TCPB to generate 300 word region proposals for each image in the ICDAR 2013 testing set. Next, we assign a positive label to word region proposals that have an IoU overlap greater than 0.5 with a ground-truth bounding box, while assigning a negative label to proposals that has an IoU overlap with any ground-truths of less than 0.5. In total, we collected 8,481 positive word region proposals and 61,419 negative word region proposals. We then evaluated the true positive (TP) rate and false positive (FP) rate of the baseline model and model employing ATC and MLRP. The results are shown in Table 1. It can be seen that the model using ATC and MLRP increase the TP rate by 3.13\% and decrease the FP rate by 0.82\%, which shows that the incorporation of more reasonable supervised and multi-level information is effective for learning more discriminative features to distinguish text from complex and diverse backgrounds.
\begin{figure}
\centering
\includegraphics[width=115mm,height=25mm]{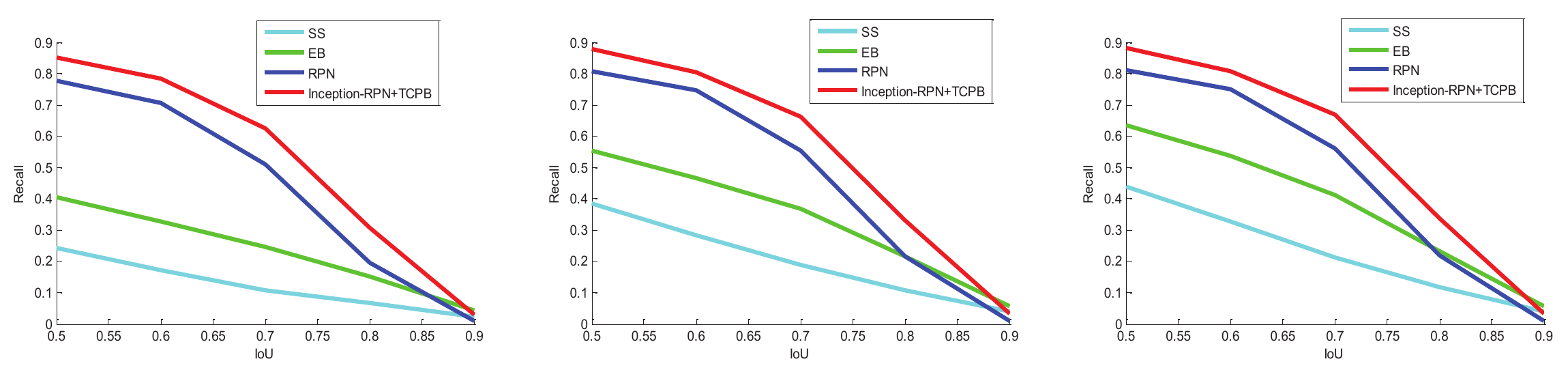}
\caption{
Recall vs. IoU overlap threshold on the ICDAR 2013 testing set. Left: 100 word region proposals. Middle: 300 word region proposals. Right: 500 word region proposals.}
\label{fig:graph}
\end{figure}

\renewcommand{\normalsize}{\fontsize{9.6pt}{11pt}\selectfont}
\begin{table}
\begin{center}
\begin{tabular}{|l|c|c|}
\hline
Model & TP(\%) & FP(\%) \\
\hline\hline
Baseline model & 85.61 & 11.20 \\ \hline
ATC+MLRP & \bf{88.74} & \bf{10.38} \\
\hline
\end{tabular}
\end{center}
\caption{Performance evaluation of ATC and MLPB based on TP and FP rate.}
\end{table}

\subsection{Experimental results on full text detection}
We evaluate the proposed DeepText detection system on the ICDAR 2011 and 2013 robust text detection benchmarks following the standard evaluation protocol of ICDAR 2011 \cite{IJDAR06} and 2013 \cite{ICDAR2013}. Our DeepText system achieves {\bf0.83} and {\bf0.85} F-measure on the ICDAR 2011 and 2013 datasets. Comparisons with recent methods on the ICDAR 2011 and 2013 benchmarks are shown in Tables 2 and 3. It is worth to note that though Sun \emph{et al.} \cite{CER} achieved superior results on the ICDAR 2011 and 2013 datasets, their method is not comparable because they used millions of additional samples for training, while we only used 4072 training samples. In the tables, we can see that our proposed approach outperforms previous results with a substantial improvement, which can be attributed to simultaneously taking high recall and precision into consideration in our system. The High performance achieved on both datasets highlights the robustness and effectiveness of our proposed approach. Further, qualitative detection results under diverse challenging conditions are shown in Fig. 4, which demonstrates that our system is capable of detecting non-uniform illumination, multiple and small regions, as well as low contrast text regions in natural images. In addition, our system takes 1.7 s for each image on average when using a single GPU K40.

\renewcommand{\normalsize}{\fontsize{9.6pt}{11pt}\selectfont}
\begin{table}
\begin{center}
\begin{tabular}{|l|c|c|c|c|}
\hline
Method & Year & Precision & Recall & F-measure \\
\hline\hline
DeepText (ours) & N/A & 0.85 & {\bf0.81} & {\bf0.83} \\ \hline
TextFlow \cite{TextFlow} & ICCV 2015 & 0.86 & 0.76 & 0.81 \\ \hline
Zhang \emph{et al.} \cite{Baixiang} & CVPR 2015 & 0.84 & 0.76 & 0.80 \\ \hline
MSERs-CNN \cite{MSER4} & ECCV 2014 & {\bf0.88} & 0.71 & 0.78 \\ \hline
Yin \emph{et al.} \cite{MSER3} & TPAMI 2014 & 0.86 & 0.68 & 0.75 \\ \hline
Faster-RCNN \cite{Faster-RCNN} & NIPS 2015 & 0.74 & 0.71 & 0.72 \\ \hline
\end{tabular}
\end{center}
\caption{Comparison with state-of-the-art methods on the ICDAR 2011 benchmark.}
\end{table}

\renewcommand{\normalsize}{\fontsize{9.6pt}{11pt}\selectfont}
\begin{table}
\begin{center}
\begin{tabular}{|l|c|c|c|c|}
\hline
Method & Year & Precision & Recall & F-measure \\
\hline\hline
DeepText (ours) & N/A & 0.87 & {\bf0.83} & {\bf0.85} \\ \hline
TextFlow \cite{TextFlow} & ICCV 2015 & 0.85 & 0.76 & 0.80 \\ \hline
Zhang \emph{et al.} \cite{Baixiang} & CVPR 2015 & 0.88 & 0.74 & 0.80 \\ \hline
Lu \emph{et al.} \cite{IJDAR} & IJDAR 2015 & {\bf0.89} & 0.70 & 0.78 \\ \hline
Neumann \emph{et al.}\cite{ICDAR} & ICDAR 2015 & 0.82 & 0.72 & 0.77 \\ \hline
FASText \cite{Fastext} & ICCV 2015 & 0.84 & 0.69 & 0.77 \\ \hline
Iwrr2014 \cite{ACCV} & ACCVW 2014 & 0.86 & 0.70 & 0.77 \\ \hline
Yin \emph{et al.} \cite{MSER3} & TPAMI 2014 & 0.88 & 0.66 & 0.76 \\ \hline
Text Spotter \cite{CVPR2012} & CVPR 2012 & 0.88 & 0.65 & 0.75 \\ \hline
Faster-RCNN \cite{Faster-RCNN} & NIPS 2015 & 0.75 & 0.71 & 0.73 \\ \hline
\end{tabular}
\end{center}
\caption{Comparison with state-of-art methods on the ICDAR 2013 benchmark.}
\end{table}

\begin{figure}
\centering
\includegraphics[width=120mm]{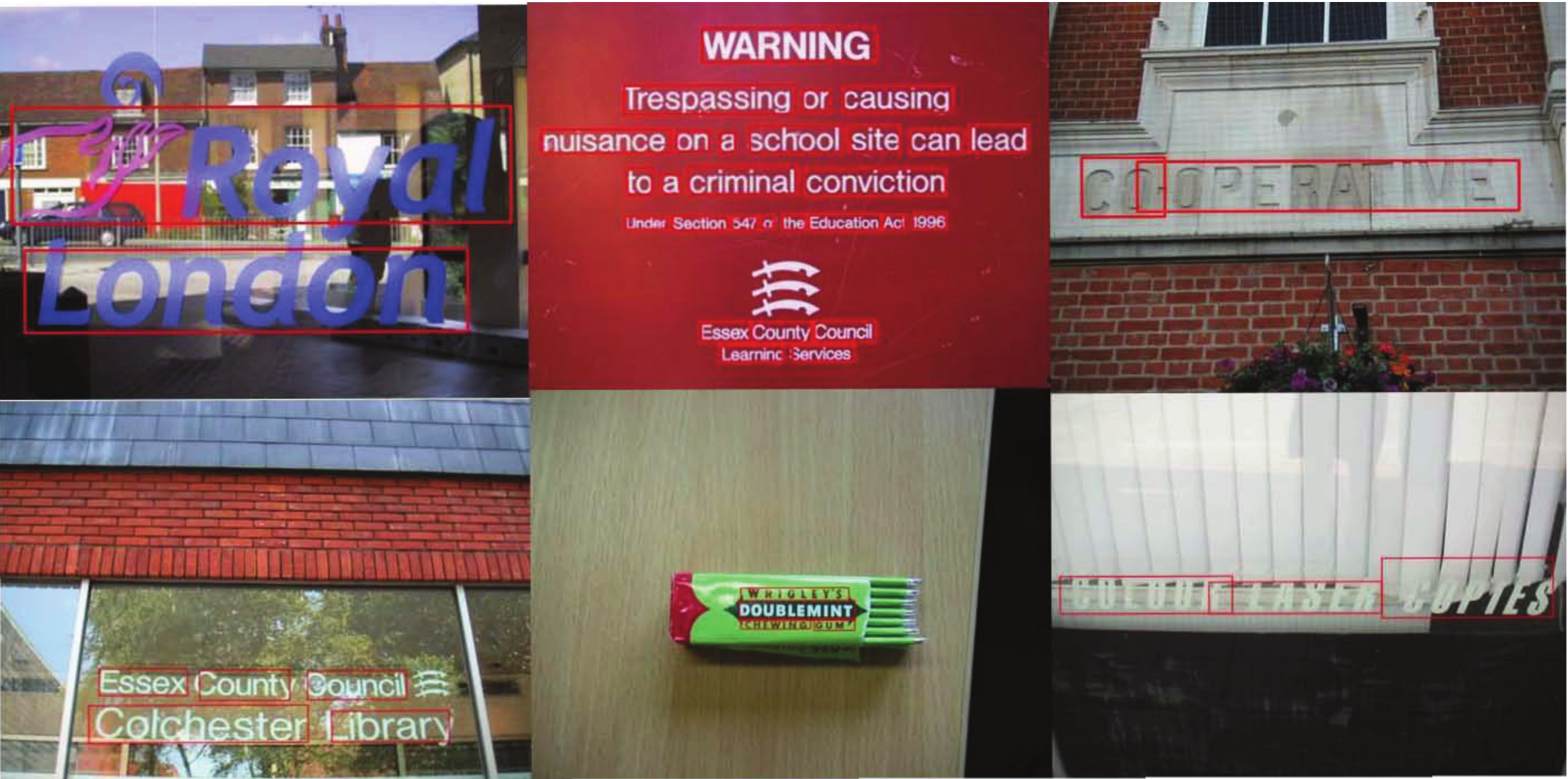}
\caption{
Example detection results of our DeepText system on the ICDAR 2011 and ICDAR 2013 benchmarks.
}
\label{fig:graph}
\end{figure}

\section{Conclusion}
\label{sec:conclusion}
In this paper, we presented a novel unified framework called DeepText for text detection in natural images with a powerful fully CNN in an end-to-end learning manner. DeepText consists of an Inception-RPN with a set of text characteristic prior bounding boxes for high quality word proposal generation and a powerful text detection network for proposal classification and accurate localization. After applying an iterative bounding box voting scheme and filtering algorithm to remove redundant boxes for each text instance, we achieve our high-performance text detection system. Experimental results show that our approach achieves state-of-the-art F-measure performance on the ICDAR 2011 and 2013 robust text detection benchmarks, substantially outperforming previous methods. We note that there is still a large room for improvement with respect to recall and precision. In future, we plan to further enhance the recall rate of the candidate word region proposals and accuracy of the proposal classification and location regression.

\bibliography{egbib}
\end{document}